\documentclass[a4paper,fleqn]{cas-sc}
\usepackage[authoryear]{natbib}

\begin{document}

\let\WriteBookmarks\relax
\def\floatpagepagefraction{1}
\def\textpagefraction{.001}
\let\printorcid\relax
\shorttitle{}
\shortauthors{T.Zhou et~al.}

\title [mode = title]{Path and Bone-Contour Regularized Unpaired MRI-to-CT Translation}                      
\author[1]{Teng Zhou}
\ead{2112305323@mail2.gdut.edu.cn}

\credit{Conceptualization, Formal analysis, Investigation, Methodology, Software, Validation, Visualization, Writing - Original Draft, Writing - Review \& Editing}

\author[2]{Jax Luo}
\ead{jluo5@mgh.harvard.edu}
\credit{Resources, Supervision, Validation, Writing - Review \& Editing}

\author[1]{Yuping Sun}
\cormark[1]
\ead{syp@gdut.edu.cn}
\credit{Conceptualization, Funding acquisition, Methodology, Project administration, Resources, Supervision, Writing - Review \& Editing}

\author[3,5]{Yiheng Tan}
\ead{y.tan@umcg.nl}
\credit{Data curation, Investigation, Visualization}

\author[3]{Shun Yao}
\cormark[1]
\ead{yaosh23@mail.sysu.edu.cn}
\credit{Data curation, Funding acquisition, Resources, Supervision, Writing - Review \& Editing}

\author[4]{Nazim Haouchine}
\ead{nhaouchine@bwh.harvard.edu}
\credit{Validation, Writing - Review \& Editing}

\author[2]{Scott Raymond}
\ead{raymons3@ccf.org}
\credit{Validation, Writing - Review \& Editing}

\address[1]{School of Computer Science and Technology, Guangdong University of Technology,
Guangzhou, China}
\address[2]{Neurological Institute, Cleveland Clinic, OH, USA}
\address[3]{Department of Neurosurgery, The First Affiliated Hospital, Sun Yat-sen University, Guangzhou, China}
\address[4]{Brigham and Women’s Hospital, Harvard Medical School, MA, USA}
\address[5]{Department of Radiology, Medical Imaging Center, University Medical Center Groningen, University of Groningen, 9712 CP Groningen, The Netherlands}

\cortext[cor1]{Corresponding author}

\begin{abstract}
Accurate MRI-to-CT translation promises the integration of complementary imaging information without the need for additional imaging sessions. Given the practical challenges associated with acquiring paired MRI and CT scans, the development of robust methods capable of leveraging unpaired datasets is essential for advancing the MRI-to-CT translation. Current unpaired MRI-to-CT translation methods, which predominantly rely on cycle consistency and contrastive learning frameworks, frequently encounter challenges in accurately translating anatomical features that are highly discernible on CT but less distinguishable on MRI, such as bone structures. This limitation renders these approaches less suitable for applications in radiation therapy, where precise bone representation is essential for accurate treatment planning. To address this challenge, we propose a path- and bone-contour regularized approach for unpaired MRI-to-CT translation. In our method, MRI and CT images are projected to a shared latent space, where the MRI-to-CT mapping is modeled as a continuous flow governed by neural ordinary differential equations. The optimal mapping is obtained by minimizing the transition path length of the flow. To enhance the accuracy of translated bone structures, we introduce a trainable neural network to generate bone contours from MRI and implement mechanisms to directly and indirectly encourage the model to focus on bone contours and their adjacent regions. Evaluations conducted on three datasets demonstrate that our method outperforms existing unpaired MRI-to-CT translation approaches, achieving lower overall error rates. Moreover, in a downstream bone segmentation task, our approach exhibits superior performance in preserving the fidelity of bone structures. Our code is available at: https://github.com/kennysyp/PaBoT.
\end{abstract}




\maketitle

\section{Introduction}

Magnetic Resonance Imaging (MRI) and Computed Tomography (CT) are foundational modalities in medical imaging, each with distinctive advantages. MRI measures signal intensity from tissue responses to a magnetic field, providing exceptional soft tissue contrast that is invaluable in diagnosing conditions such as tumors and neurological disorders. In contrast, CT quantifies tissue radiodensity in Hounsfield Units, making it highly effective for visualizing bone structures, identifying acute lesions, and calculating radiation dose in radiotherapy. The integration of MRI and CT, conventionally by non-rigid image registration, is particularly advantageous in diagnosis and treatment planning. However, acquiring both MRI and CT images is time-consuming and costly, and CT scanning also involves ionizing radiation, posing risks to vulnerable populations, including pregnant women and children. Additionally, the inherent misalignment between the MRI and CT complicates the registration process, leading to potential errors that can compromise diagnostic accuracy \citep{CycleGAN1, CycleGAN2}. Addressing these challenges necessitates the development of efficient and precise algorithms capable of translating MRI images into corresponding CT counterparts. Such advancements hold potential to alleviate patient burden, streamline clinical workflows, and optimize treatment planning \citep{cDDPM1}. 

\begin{figure}[!t]
\centerline{\includegraphics[width=0.9\columnwidth]{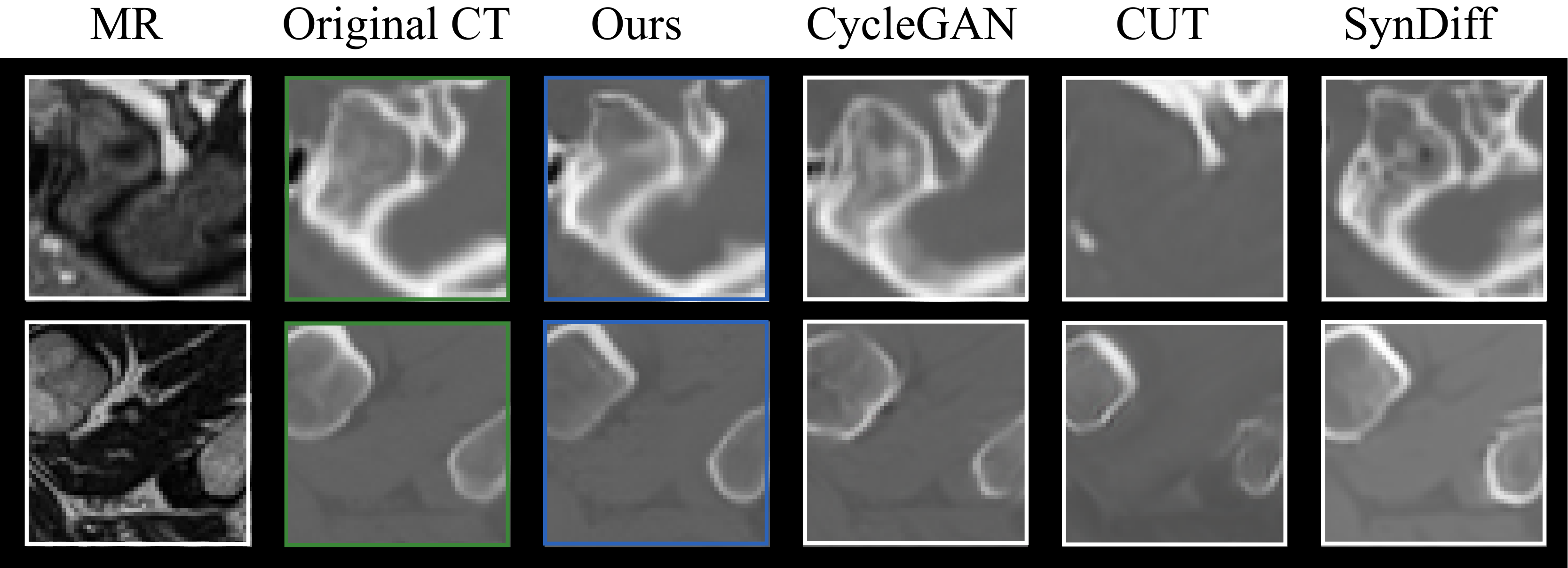}}
\caption{Qualitative comparison of the proposed approach with mainstream unpaired MRI-to-CT translation methods. The top and bottom rows depict zoomed-out regions from the head and the pelvic datasets, respectively. Our method demonstrates superior bone fidelity compared to competing approaches.}
\label{fig1}
\end{figure}

Early methods for MRI-to-CT translation primarily relied on image registration and atlas-based techniques \citep{burgos2015robust,sjolund2015generating}, whose performance was limited by registration errors. Recent advancements have seen a transition toward the use of generative models such as generative adversarial networks (GAN) \citep{GAN_theory1} and variational autoencoders (VAE) \citep{VAE}. GAN and VAE encode the MRI into a latent space and then decode it to create a CT-like image. However, they do it in a single-step sampling fashion, inherently limiting the reliability of the translation and being prone to issues like premature convergence and mode collapse \citep{DDPM_theory2, DDPM_theory1, DDPM2025}. Lately, denoising diffusion probabilistic models (DDPMs) have gained popularity in MRI to CT translation. In the forward process, DDPMs begin with a CT image and iteratively add Gaussian noise in small increments. After a large number of steps, the CT becomes nearly indistinguishable from random noise. In the reverse process, the model progressively predicts the denoised version of the CT image, conditioned on the input MRI. Since DDPMs operate as continuous-time models, they typically generate sharper and more anatomically accurate CT images compared to single-step GANs or VAEs, albeit with the drawback of slower inference times. It is noteworthy that the training of paired-data-based GANs, VAEs, and DDPMs requires precisely aligned MR-CT datasets to produce high-quality synthetic CT outputs \citep{pix2pix, cGAN1, cGAN3, ResViT, cDDPM1, cGAN6, Bridge1,cDDPM3, cGAN4, cDDPM2}. Consequently, the clinical applicability of these paired methods remains constrained by the practical challenges in acquiring such well-registered MRI and CT data pairs.

Unpaired MRI-to-CT translation is a fundamentally ill-posed problem due to the potentially infinite number of mappings between the MRI and CT domains. To address this, researchers have introduced cycle consistency, which assumes a one-to-one correspondence between the MRI and CT domains. Prominent methods based on this principle include CycleGAN \citep{CycleGAN1, CycleGAN_theory, CycleGAN4, CycleGAN3, CycleGAN5}, SynDiff \citep{SynDiff}, and FDDM \citep{FDDM}. However, the one-to-one assumption can be overly restrictive, particularly when MRI and CT data contain domain-specific information that is not directly transferable. Another widely used approach for unpaired MRI-to-CT translation is Contrastive Learning (CL) \citep{CUT_theory}. CL aims to maximize the mutual information between corresponding patches in MRI and CT images by employing loss functions to associate matching patches while disassociating non-corresponding ones \citep{CUT1,CUT2}. Despite its promise, the efficacy of CL is highly sensitive to the selection of sample pairs. This issue is particularly pronounced in MRI-to-CT translation, where the visual characteristics of an MRI patch could differ markedly from its corresponding CT patch, leading to suboptimal outcomes. Another notable limitation of existing unpaired MRI-to-CT translation methods is their inability to preserve bone boundary contours in the translated CT images accurately. This challenge arises from the inherently poor visibility of bones in MRI, leading to deviations in the translated CT's bone boundary delineation compared to the reference CT. These inaccuracies can substantially compromise the precision of dose calculations in radiotherapy treatment planning, thereby undermining the clinical applicability and reliability of these methods.

In this paper, we introduce a path- and bone-contour regularized approach for unpaired MRI-to-CT translation. Departing from the cycle-consistency and contrastive learning strategies, we project MRI and CT images into a shared latent space and model the MRI-to-CT mapping as a continuous flow governed by neural ordinary differential equations. The optimal mapping is obtained by minimizing the transition path length of the flow, facilitating training via a straightforward least-squares optimization. Our approach mitigates the instability typically observed in one-step methods like GANs and delivers substantially faster inference times compared to continuous-time models like DDPMs. To enhance the accuracy of translated bone structures, we introduce a trainable neural network to generate bone contours from MRI and implement mechanisms to directly and indirectly encourage the model to focus on bone contours and their adjacent regions, even in the absence of paired CT images in the training dataset. Extensive experimental validation conducted on three distinct public datasets has demonstrated that our approach provides better image quality and fidelity for MRI-to-CT translation compared to competing GAN, diffusion, and CL models. Moreover, the effectiveness of the bone-contour regularization technique is validated through a downstream bone segmentation task, where our method achieves significantly higher DICE scores than all competing methods. Fig.~\ref{fig1} illustrates qualitative results for MRI-to-CT translation, highlighting the enhanced bone fidelity of our approach. 

\section{Related Work}

\subsection{Unpaired MRI-to-CT Translation}

The difficulty of acquiring strictly aligned MRI and CT datasets has spurred a growing interest in unpaired image translation techniques. Current approaches for unpaired MRI-to-CT translation predominantly leverage two key strategies: cycle consistency and contrastive learning \citep{review1,review2}.

Cycle consistency addresses the challenge of infinite possible mappings in unpaired MRI-to-CT translation by enforcing a one-to-one correspondence between the two image domains. A seminal method employing this strategy is CycleGAN \citep{CycleGAN1,CycleGAN_theory,CycleGAN4,CycleGAN3,CycleGAN5,CycleGAN2}. In CycleGAN, an image $x$ from the MRI domain is translated to the CT domain as $y_{G_1}$ using a generator $G_1$ and subsequently mapped back to the MRI domain via another generator $G_2$, producing a synthetic MRI $x_{G_2}$ that closely matches the original input $x$. This bidirectional consistency constraint enables CycleGAN to learn realistic and reversible mappings, establishing it as a robust framework for unpaired MRI-to-CT translation. However, the generators of CycleGAN utilize a single-step encoder-decoder architecture, which is prone to challenges such as premature convergence and mode collapse. Recently, a more advanced generator Denoising Diffusion Probabilistic Model (DDPM \citep{DDPM_theory2,DDPM_theory1}) $G_\mathrm{DDPM}$ has gained traction in MRI to CT translation \citep{cDDPM1,cDDPM2,cDDPM3, cDDPM-latent}. DDPM is a generative model that has a forward and reverse diffusion process. In the forward process, DDPM begins with a CT image and iteratively adds Gaussian noise in small increments. After a large number of steps, the CT becomes nearly indistinguishable from random noise. In the reverse process, the model progressively predicts the denoised version of the CT image, conditioned on the input MRI. Since DDPM functions as a continuous-time generator, DDPM-based cycle-consistent image translation models, such as SynDiff \citep{SynDiff}, are generally more reliable and output sharper CT images compared to single-step CycleGAN, at the cost of slower inference times. Despite its advantages, cycle consistency has inherent limitations. Notably, cycle consistency enforces a strict one-to-one correspondence, even in cases where such an assumption may not be justified. For example, when mapping back from $y_x$ to $x$, it requires the synthetic MRI $x_y$ to reconstruct the original $x$. However, the bright bone signal in $y_x$, a CT-specific feature, has no direct counterpart in MRI, where bone and its neighboring region appear as a dark void. Consequently, the model may introduce artifacts or fail to recover $x$ accurately.

\begin{figure}[!t]
    \centering
    \includegraphics[width=1\textwidth]{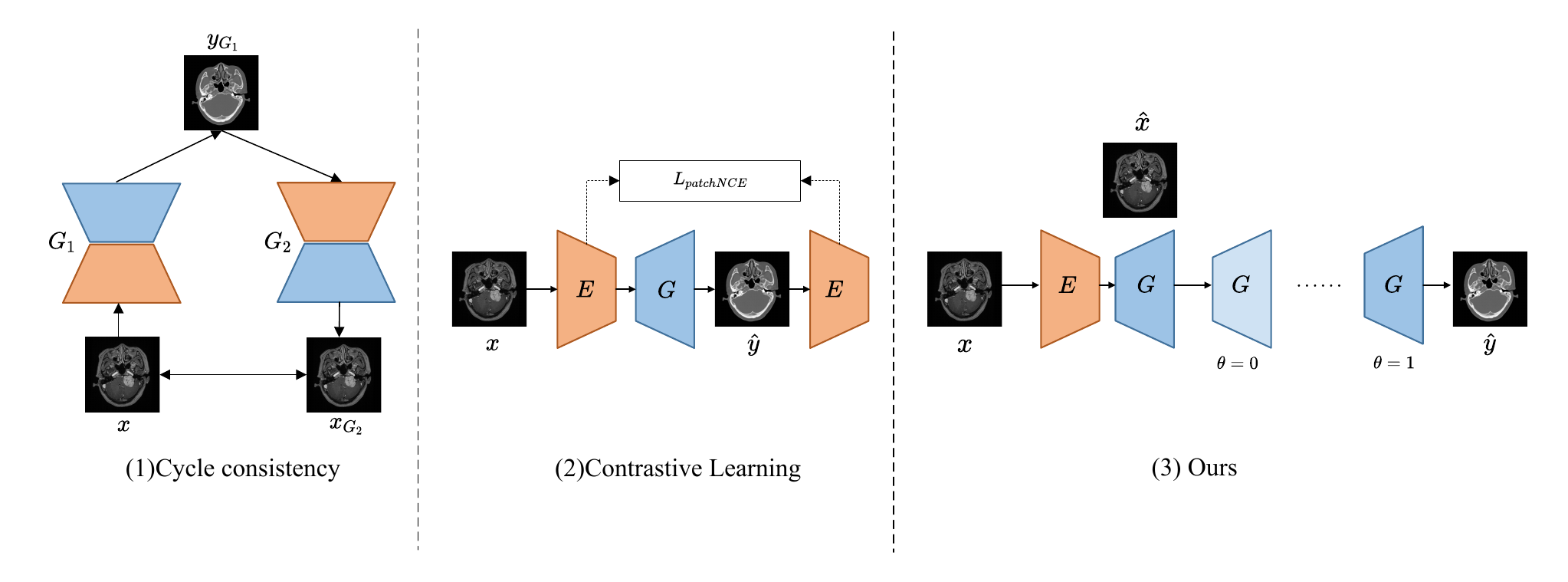}
    \caption{A schematic overview of Cycle Consistency, Contrastive Learning, and the proposed framework. Here, $x$ and $y$ represent the MRI and CT images, respectively. (1) $y_{G_1}$ is the synthetic CT, while $x_{G_2}$ is the synthetic MRI to ensure cycle consistency. $G_1$ and $G_2$ denote respective generators. (2) $E$ and $D$ denote the encoder and decoder. The latent representation of the CT image is given by $\hat{y}$, and the contrastive loss is denoted by $\mathit{L}_{\mathit{patch}\mathrm{NCE}}$. (3) $\hat{x}$ and $\hat{y}$ correspond to the latent representation of MRI and CT images. $E$ and $G$ refer to the encoder and decoder modules, and $\theta\in[0,1]$ is a continuous time variable.}
    \label{fig2}
\end{figure}

The Contrastive Learning (CL) framework in MRI-to-CT image translation is designed to learn a shared feature representation that aligns corresponding anatomical or structural information between MRI and CT domains. This approach \citep{CUT1,CUT4,CUT2} decomposes the MRI and CT images into corresponding patches, and leverages the concept of contrastive loss $\mathit{L}_{\mathit{patch}\mathrm{NCE}}$ to encourage positive pairs of selected patch samples to have similar embeddings while maximizing the difference between negative pairs. The efficacy of CL is highly sensitive to the selection of sample patch pairs. This issue is particularly pronounced in MRI-to-CT translation, where the visual characteristics of an MRI patch could differ markedly from its corresponding CT patch, leading to suboptimal outcomes. Grounded in the theory of optimal transport, the proposed approach projects MRI and CT images into a shared latent space and models the MRI-to-CT translation as a continuous flow process, parameterized by a neural network $G$ over $\theta\in[0,1]$. $G$ starts from the MRI image's latent representation at $\theta=0$ and maps it to the corresponding CT at $\theta=1$. The optimal mapping is derived by minimizing the transition path length of the flow. The proposed approach facilitates efficient training through a straightforward least-squares optimization and effectively mitigates the limitations of Cycle Consistency and Contrastive Learning. Fig.~\ref{fig2} presents a simplified overview of Cycle Consistency, Contrastive Learning, and the proposed approach.

\subsection{Bone-contour Regularization}

Utilizing edge information as a regularization has been an effective way to enhance the anatomical fidelity for image translation. For example, Ea-GANs \citep{Ea-GANs} leverages a Sobel filter to produce an edge map that accentuates structural boundaries in MRI images, including tissue interfaces and anatomical contours. This edge information is integrated into both the generator and discriminator, facilitating the adversarial learning of edge similarity to improve the accuracy and fidelity of cross-modality MRI image translation. FDDM \citep{FDDM} also applies a Sobel filter to extract edges from unpaired MR and CT images. The extracted MRI edges are converted into corresponding CT edges using a variational autoencoder (VAE). These transformed edges are then used as conditional inputs into the diffusion model, ensuring high boundary fidelity for MRI to CT image translation. ContourDiff \citep{Contour_diff} utilizes a Canny edge detection filter to extract contour representations, which are integrated with the diffusion model input at each denoising step. This integration is performed in the image space to ensure that the CT-to-MRI translation strictly adheres to the contour guidance. All the aforementioned methods utilize gradient-based techniques to extract edge information. While these approaches are effective for CT images, they are less suitable for MRI due to the low gradient contrast between bone and surrounding structures, such as air, which hampers accurate bone boundary detection.

In this study, instead of applying filters to generate a global edge map for the entire MRI image, we introduce a trainable contour generator specifically designed to extract bone contours. Furthermore, we incorporate mechanisms that explicitly and implicitly direct the model’s focus toward bone contours and their surrounding regions, promoting a more accurate and faithful representation of bone structures in the translated CT.

\section{Method}

In this section, we begin by formulating the MRI-to-CT translation problem as the task of finding an optimal transition path within the latent space. We then present the intuition behind our approach and provide a detailed explanation of the path and bone-contour regularization techniques. Finally, we combine all ingredients and introduce the full objective function of the network.

\subsection{Problem Definition}
Consider a random vector $X \in \mathbb{R}^3$ representing MRI images with marginal distribution $\pi_0$, and a random vector $Y\in \mathbb{R}^3$ representing CT images with marginal $\pi_1$. The objective is to determine an optimal mapping from MRI to CT, which correspond to inferring the true joint distribution $P_{X,Y} = P_{Y \mid X}P_{X}$. Drawing on prior work \citep{Bridge1,Senta}, we estimate $P_{X,Y}$ in a shared latent space $\mathcal{Z}$ rather than the image space, abstracting complex image-level differences into latent feature distributions for more effective comparison and alignment.

\begin{figure}[!t]
\centerline{\includegraphics[width=0.9\columnwidth]{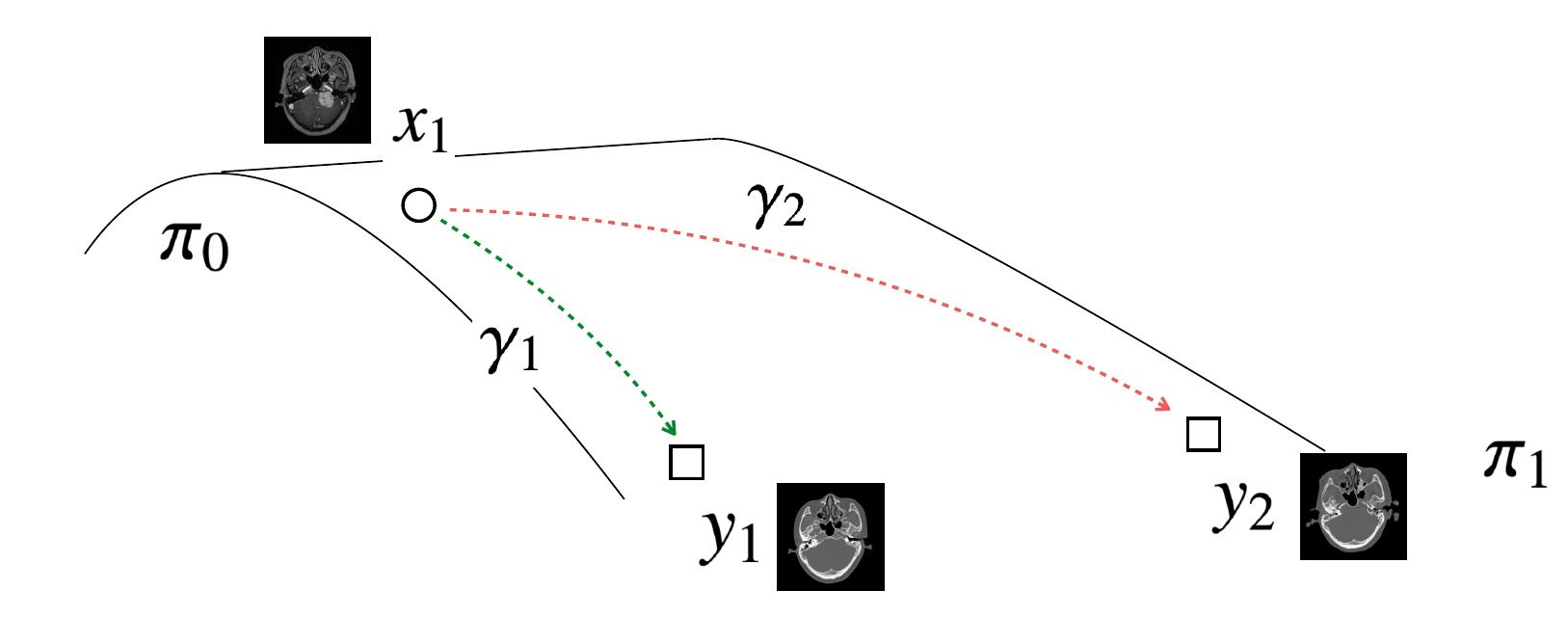}}
\caption{Illustration of the intuition of the path regularization. Translating an MRI image $x_1$ to its true CT counterpart $y_1$ should require fewer cumulative changes and less transformation effort than mapping it to an alternative CT point $y_2$, resulting in a shorter transition path length.}
\label{fig3}
\end{figure}
 
\textbf{Latent Space Embedding.} Let $x\sim \pi_{0}$ and $y\sim \pi_{1}$ denote arbitrary MRI and CT images, we first employ a shared encoder $E$ to extract their latent feature representations: 
$$
z_x=E(x), \quad z_y=E(y).
$$
To ensure $z_x$ and $z_y$ both reside in a common latent space $\mathcal{Z}$, we match their distribution, $q\left(z_x\right)$ and $q\left(z_y\right)$, using Kullback-Leibler (KL) divergence:
\begin{equation}
\mathcal{L}_{\mathrm{kl}}=\mathrm{KL}\left(q\left(z_x\right) \| p(z)\right)+\mathrm{KL}\left(q\left(z_y\right) \| p(z)\right),
\end{equation}
where $p(z)$ is an isotropic Gaussian $\mathcal{N}(0, I)$ prior distribution for $z \in \mathcal{Z}$.

Subsequently, we train a decoder $G$ to reconstruct $x$ and $y$ as latent space embeddings $\hat{x}=G(z_x)$ and $\hat{y}=G(z_y)$ by minimizing the following losses:
\begin{equation} \label{eq1}
    \begin{aligned}
        \mathcal{L}_{\text {x}}=\mathbb{E}_{x \sim \pi_{0}}\left\|G\left(z_x\right)-x\right\|_1, \\
        \mathcal{L}_{\text {y}}=\mathbb{E}_{y \sim \pi_{1}}\left\|G\left(z_y\right)-y\right\|_1.
    \end{aligned}
\end{equation}

\textbf{Transition Path} The translation of MRI to CT in the latent space is formulated as a transport map $T: \mathbb{R}^3 \rightarrow \mathbb{R}^3$, such that $\hat{y}:= T(\hat{x})$. Here, the pair $(\hat{x},\hat{y})$  constitute a transport plan coupling $\pi_0$ and $\pi_1$, describing a mechanism to connect samples across these distributions. To identify the optimal transport map $T^*$, we require the optimal transport plan $(\hat{x},\hat{y}^*)$, where $\hat{y}^*$ is the true CT counterpart to $\hat{x}$. However, for unpaired MRI to CT translation, $\hat{y}^*$ is not directly accessible. 

To address this, we model the transport plan as a continuous flow governed by neural ordinary differential equations (ODEs). For any transport plan (x,y), it induces a flow
\begin{equation}
\mathrm{d}\hat{x}_\theta=v\left(\hat{x}_\theta, \theta\right) \mathrm{d} \theta,
\end{equation}
a continuous-time process over $\theta\in[0,1]$ that converts $\hat{x}$ to $\hat{y}$. Here, $v: \mathbb{R}^3 \rightarrow \mathbb{R}^3$ represents a velocity that steers the flow from $\hat{x}$ to $\hat{y}$, yielding a transition path $\gamma:[\hat{x},\hat{y}]$ by solving:
\begin{equation}
\min_{v}\int_{0}^{1} \mathbb{E}\left[\left\|(\hat{y} - \hat{x}) - v(\hat{x}_{\theta}, \theta) \right\|^{2}\right] \mathrm{d} \theta, 
\end{equation}
where $ \quad \hat{x}_{\theta} = \theta \hat{y} + (1-\theta)\hat{x}$ denotes the linear interpolation between $\hat{x}$ and $\hat{y}$. 

In practice, we parameterize the transition with the neural network $G$, which also serves as the decoder for reconstructing $\hat{x}$ and $\hat{y}$. 
The transition process begins at $\theta=0$ with the latent MRI representation $\hat{x}$, and $G(z_x,\theta)$ is trained such that, upon convergence at $\theta=1$, it produces the translated CT image $\hat{y}$.

In other words, any transport plan $(\hat{x},\hat{y})$ can be expressed as 
\begin{equation}
\hat{x}=G(z_x, 0),  \hat{y}=G(z_x, 1).
\end{equation}
By varying $\theta$ from 0 to 1, the trajectory of all intermediate outputs of $G(z_x,\theta)$ form a transition path:
\begin{equation}
    \gamma:[\hat{x},\hat{y}]=\gamma:[0,1] \rightarrow G\left(z_x, \theta\right).
\end{equation}
Hence, each transport plan $(\hat{x},\hat{y})$ uniquely defines a transition path $\gamma$. The estimation of the optimal MRI-to-CT mapping $T^*$ is thereby transformed into the determination of the optimal transition path $\gamma^*$. In the next subsection, we introduce a regularization term for $\gamma$ to aid in identifying the optimal transport plan $(\hat{x},\hat{y})$.

\subsection{Path Regularization}

In unpaired MRI to CT translation, without additional constraints, there are infinite many transport map $T$ between $\pi_0$ and $\pi_1$, leading to an infinite amount of transition path $\gamma$s. To tackle this ill-posed problem, we adopt the shortest path assumption: The optimal transition path $\gamma^*$, corresponding to the optimal transport plan $(\hat{x},\hat{y}^*)$, is assumed to be the shortest among all possible paths on a hypothetical manifold $\mathcal{M}$ connecting the two domains \citep{Senta,Bridge1}. 

\begin{figure}[!t]
    \centering
    \includegraphics[width=1\textwidth]{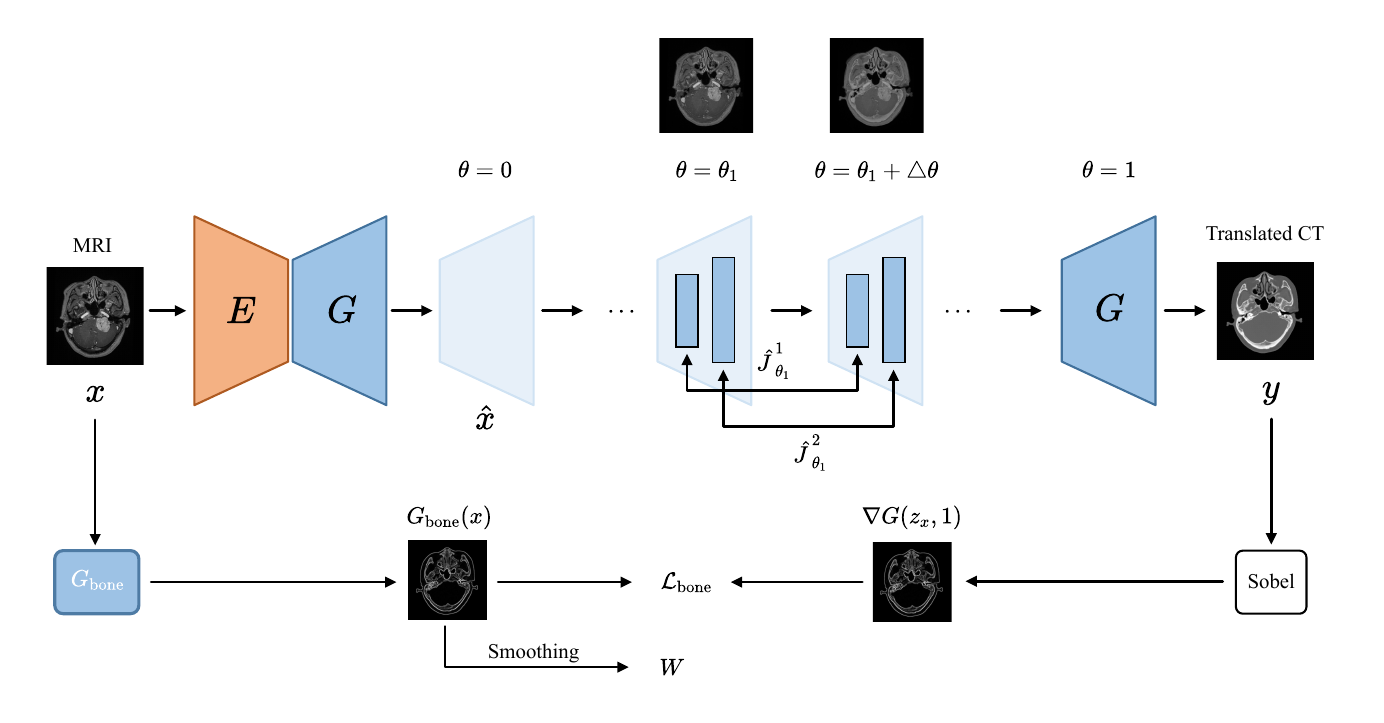}
    \caption{A more detailed illustration of the proposed framework, with particular emphasis on the layer-wise path-length regularization and bone-contour regularization components.}
    \label{fig4}
\end{figure}

This assumption stems from the principle that translating an MRI image $x_1$ to its true CT counterpart $y_1$ should be a more direct process than mapping it to alternative CT points. The simplicity of this transformation arises from the fact that $x_1$ and $y_1$, though situated in distinct domains, are hypothesized to share a common latent representation $z$, as $x_1 \approx G(z_1,0)$ and $y_1 \approx G(z_1,1)$, which captures the essential semantic features that persist across both MRI and CT domains. As illustrated in Fig.~\ref{fig3}, $y_2$ represents an alternative CT image that is not $y_1$. Since $y_2$ is generated from a different latent representation, e.g., $y_2\approx G(z_2,1)$ where $z_2 \neq z_1$, the semantic overlap between $x_1$ and $y_2$ is reduced, this mismatch implies that mapping $x_1$ to $y_2$ requires not only a domain shift but also an alteration of the latent representation, from $z_1$ to $z_2$, potentially discarding critical semantic features. Since the length of the path is a measure of the cumulative change or effort required to transform an MRI image into CT, the transition path from $x_1$ to $y_2$, $\gamma_2:[G(z_1,0),G(z_2,1)]$, likely involves a longer trajectory on $\mathcal{M}$ than $\gamma_1:[G(z_1,0),G(z_1,1)]$ of $x_1$ to $y_1$, as it must traverse additional regions of the manifold to accommodate this latent shift. Given that a shorter transition path more effectively captures semantic features in MRI, we introduce a constraint to regulate its length.

\textbf{Path Length} As described in Section 3.1, we model the MRI to CT translation as a continuous flow on $\mathcal{M}$, transforming $\hat{x}$ into $\hat{y}$ over the parameter interval $\theta \in [0,1]$. This defines a transition path $\gamma:[\hat{x},\hat{y}]$. The length of $\gamma$, as a curve on $\mathcal{M}$, is given by
\begin{equation}
\ell(\gamma)=\int_{\hat{x}}^{\hat{y}}\left\|\frac{d}{d \theta}\gamma\right\| d \theta.
\end{equation}

Given that $\gamma$ is parameterized as $\gamma:[0,1] \rightarrow G\left(z_x, \theta\right)$, Eq.(7) can be reformulated as 
\begin{equation}
\ell(\gamma)=\int_0^1\left\|\frac{d}{d\theta}G(z_x,\theta)\right\| d \theta.
\end{equation}
Since $G(z_x,\theta)$ is a neural network, its derivative with respect to $\theta$, $\frac{d}{d\theta}G(z_x,\theta)$, corresponds to its Jacobian matrix, namely $J_\theta$. Consequently, regularizing the path length reduces to penalizing:
\begin{equation}
\ell(\gamma)=\int_0^1\left\|J_\theta\right\| d \theta.
\end{equation}
Due to the computational complexity of directly computing $J_\theta$, we approximate it using the central finite difference method.
\begin{equation}
\hat{J_\theta}=\frac{G\left(z_x, \theta+\frac{h}{2}\right)-G\left(z_x, \theta-\frac{h}{2}\right)}{h},
\end{equation}
where $h$ is a hyperparameter that controls the level of detail or precision in the Jacobian estimation. A smaller $h$ provides a finer, more precise approximation, while a larger $h$ results in a coarser estimate. In practice, $h$ is randomly sampled from the uniform distribution $U(0.1,0.2)$.

Given that $G$ is a multi-layered network, it is beneficial to regularize the path length at each layer separately. Let $G$ consists of $K$ layers, and denote the Jacobian of the $k$th layer as $\hat{J^k_\theta}$. Incorporating this into the formulation, the final path regularization loss is given by:
 \begin{equation}
\mathcal{L}_{\text{path}} = \mathbb{E}_{\hat{x} \sim \pi_0} \mathbb{E}_{\theta \sim U(0,1)} \left[ \frac{1}{K} \sum_{k=1}^K \left\| \hat{J}_\theta^k \right\|^2 \right].
\end{equation}

Fig.~\ref{fig4} illustrates the concept of layer-wise path length regularization for a two-layer $G$ over the time interval $[\theta_1,\theta_1+\Delta\theta]$.

\subsection{Bone-contour Regularization}

To enhance the fidelity of translated bone structures, we encourage the network to focus more on bone boundaries. To achieve this, we introduce a bone-contour regularization loss, $\mathcal{L}_{bone}$, which directly penalizes discrepancies between the bone contours of the original MRI $x$ and the translated CT $G(z_x,1)$. 

Specifically, we incorporate a trainable neural network $G_{bone}$ to generate bone contours from $x$, while the Sobel filter, $\nabla$, is applied to detect edges in $G(z_x,1)$. The bone-contour regularization loss is defined as: 
\begin{equation}
    \mathcal{L}_{\mathrm{bone}} = \left \| \nabla G(z_x, 1) - G_{\text{bone}}(x) \right \|_{1}.
\end{equation}
It's important to note that $G_{bone}$ and $\nabla$ only operate at the start and endpoint of the transition path, respectively. Moreover, $G_{bone}$ is excluded from the backpropagation during the training of $G$.

Furthermore, to improve the accuracy of regions adjacent to bone contours, we apply bilateral filtering to smooth $G_{\text{bone}}(x)$, yielding an attention map $W$. Let $\alpha$ be a scaling factor that modulates the influence of $W$. During the training of $G$, we emphasize bone contours and their neighboring voxels by incorporating $W$ into the path regularization loss, which is reformulated as
\begin{equation}
\mathcal{L}^*_{\text{path}} = \mathbb{E}_{\hat{x} \sim \pi_0} \mathbb{E}_{\theta \sim U(0,1)} \left[ \frac{1}{K} \sum_{k=1}^K (1+\alpha W)\left\| \hat{J}_\theta^k \right\|^2 \right].
\end{equation}
Fig.~\ref{fig4} provides a concise illustration of how to calculate $\mathcal{L}_\mathrm{bone}$ and $W$.

\subsection{Overall Loss Function}

In summary, our goal is to reconstruct the MRI $x$ and CT $y$, translate $x$ to closely approximate its corresponding CT in $\pi_1$, and simultaneously minimize both the transition path length and the bone-contour discrepancy. 

To quantify image similarity, we employ a discriminator $D$, trained in an adversarial manner to distinguish $G(z_x, 1)$ from $y$. The adversarial loss is formulated as 
\begin{equation}
\mathcal{L}_{\mathrm{GAN}}=\mathbb{E}_{\hat{x} \sim \pi_0, y \sim \pi_1}\left[(D(G(z_x, 1))-1)^2+D\left(y\right)^2\right].
\end{equation}
Let $\lambda_1,\lambda_2,\lambda_3$ and $\lambda_4$ be hyperparameters that balance the different loss terms. The overall loss function for training the network is given by:
\begin{equation}
    \mathcal{L}_{\text {full }} = \lambda_1(\mathcal{L}_x+\mathcal{L}_y) + \lambda_2 \mathcal{L}_{\text{GAN}} + \lambda_{3} \mathcal{L}^*_{path} + \lambda_{\text{4}} \mathcal{L}_{\text{bone}}.
\end{equation}

\section{Experiments}
\subsection{Datasets and Preprocessing}
To evaluate the performance of our model, we conducted experiments using three distinct datasets: a proprietary head MRI-CT dataset provided by the First Affiliated Hospital of Sun Yat-sen University, the multimodal pelvic MRI-CT dataset \citep{nyholm2018mr}, and the SynthRAD2023 dataset \citep{thummerer2023synthrad2023}.

\subsubsection{Head Dataset}
The proprietary head MRI-CT dataset consisted of T1-weighted contrast-enhanced (T1-CE) MRI and CT images from 60 preoperative neurosurgical patients, which included 20 cases each of meningiomas, vestibular schwannomas, and cerebrovascular diseases. All images were resampled to a uniform voxel spacing of 1 $\times$ 1 $\times$ 1 mm$^3$ and reconstructed at a resolution of 256 $\times$ 256 pixels. MR volumes were corrected using N4 bias field and registered to the corresponding CT images. CT images were denoised to reduce imaging artifacts \citep{cGAN2}. This dataset was split into three groups: 40 patients for training, 5 for validation, and 15 for testing. 

All research procedures in this study were approved by the Ethics Committee of Clinical Research and Animal Trials of the First Affiliated Hospital of Sun Yat-sen University ([2022]057).

\subsubsection{Pelvic Dataset}
The Gold Atlas male pelvis dataset \citep{nyholm2018mr} contains co-registered T2-weighted (T2w) MRI and CT image pairs from 19 patients. We selected T2w MRI and corresponding CT images from 15 healthy people and divided them into three sets: 9 for training, 2 for validation, and 4 for testing.

\subsubsection{SynthRAD2023 Dataset}
The SynthRAD2023 dataset \citep{thummerer2023synthrad2023}, which includes CT, CBCT, and MRI scans from patients receiving brain and pelvic radiotherapy, was incorporated into our study. Specifically, we utilized the brain data from this dataset, partitioning it into three subsets: 40 patients for training, 5 for validation, and 15 for testing. Preprocessing and image registration were performed in accordance with the official guidelines.

\subsection{Implementation Details}  
The proposed method was implemented in PyTorch and executed on an NVIDIA RTX 4090 GPU with 24 GB of video memory. The Adam optimizer was utilized for training, with hyperparameters $\beta_{1} = 0.5$ and $\beta_{2} = 0.9$. The contour generator was based on the resnet\_6blocks architecture adopted from the generator framework in \citep{CUT_theory}. Key hyperparameter values were set as follows: the scaling factor $\alpha$ was set to 1, and the number of cross-validations was set to 100. The learning rate was configured to $2.0 \times 10^{-4}$, while the weights for contour loss, path loss, reconstruction loss, and adversarial loss were set to 5.0, 0.1, 5.0, and 1.0, respectively.

\begin{figure*}[!t]
    \centering
    \includegraphics[width=1.0\textwidth]{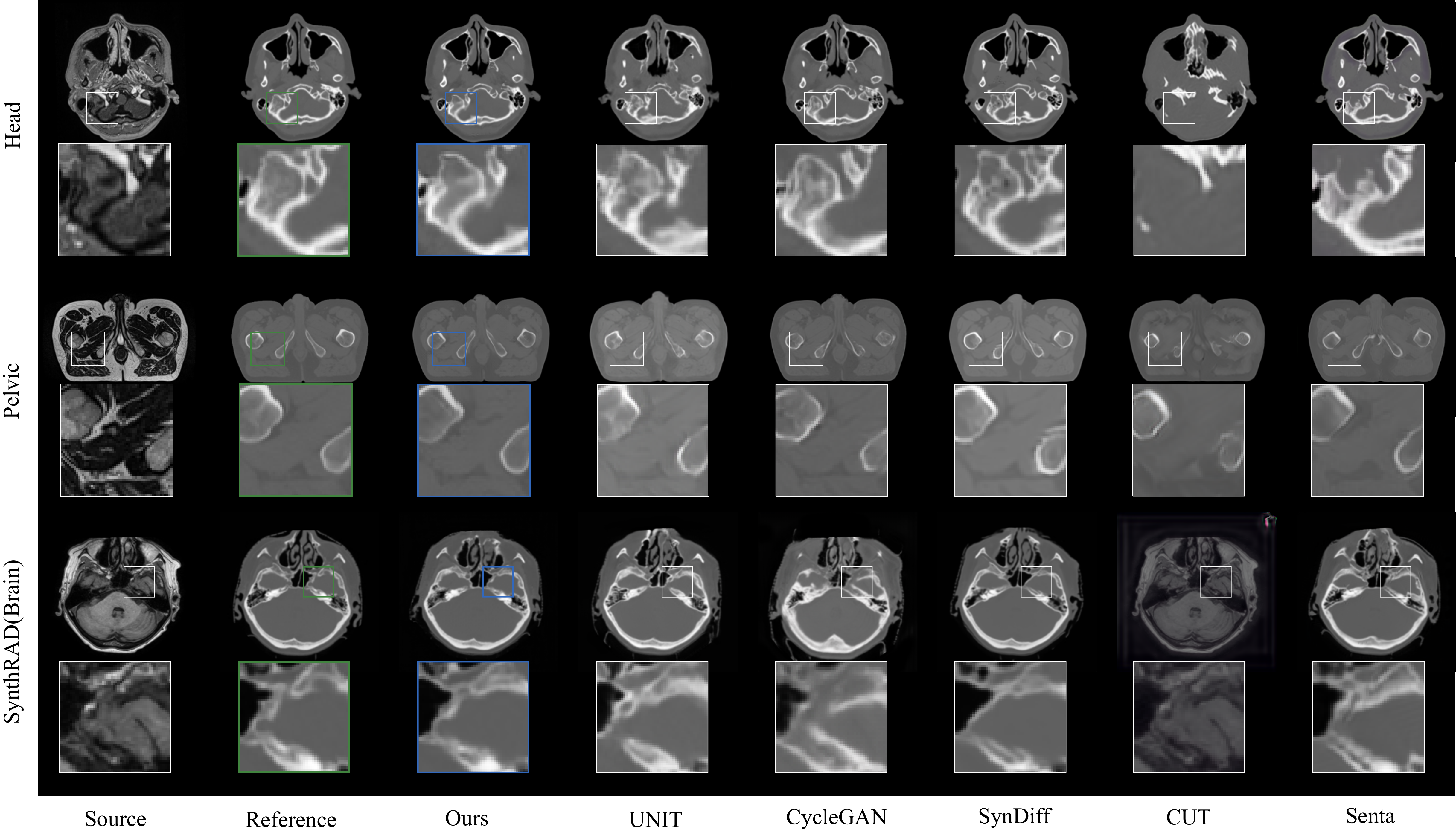}
    \caption{Qualitative comparison for all translated CTs. From left to right, the source MRI, the reference CT, and all translated CTs. From top to bottom, one example is presented for each of the Head, Pelvic, and SynthRAD datasets.}
    \label{fig5}
\end{figure*}

\subsection{Competing Methods}  

We evaluate the proposed method against five benchmark unpaired image translation techniques, namely UNIT \citep{UNIT_theory}, CycleGAN \citep{CycleGAN_theory}, SynDiff \citep{SynDiff}, CUT \citep{CUT_theory}, and Senta \citep{Senta}.

\textbf{UNIT}: UNIT is a generative model \citep{UNIT_theory} based on variational autoencoders \citep{VAE}. We trained UNIT using cross-validation over 100 epochs with a learning rate of $1.0 \times 10^{-4}$. The weights for the cycle consistency, adversarial, and reconstruction losses were configured to 10, 1, and 10, respectively.

\textbf{CycleGAN}: CycleGAN is a benchmark in unpaired MRI-to-CT translation \citep{CycleGAN_theory}. Our CycleGAN model was trained for more than 200 epochs. The learning rate began at $2.0 \times 10^{-4}$ and linearly decayed to zero after 100 epochs. The adversarial loss was weighted by 1, and the cycle consistency loss by 100.

\textbf{SynDiff}: We trained SynDiff \citep{SynDiff}, a diffusion-based image translation model with cycle consistency, for 50 epochs using cross-validation. Key hyperparameters included $T=1000$, $k=250$, and an adversarial loss weight of 1, with a learning rate initialized at $1.5 \times 10^{-4}$.

\textbf{CUT}: CUT is a contrastive learning-based method for unsupervised image translation \citep{CUT_theory}. We performed cross-validation over 200 epochs. The learning rate began at $2.0 \times 10^{-4}$ for the first 100 epochs and then linearly decreased to zero.

\textbf{Senta}: Senta is a recent ordinary differential equations-based unpaired image translation approach \citep{Senta}. Cross-validation was carried out over 200 epochs, with the learning rate initially set to $2.0 \times 10^{-4}$ for the first 100 epochs, then decaying linearly to zero. The path transfer loss weight was set to $\lambda_{path}=0.10$.

\subsection{Results}  
\subsubsection{Comparison with State-of-the-Art Methods}

\begin{table}[!t]
    \centering
    \caption{Performance of MRI-CT Translation on Various Datasets. PSNR (dB), SSIM (\%), and LPIPS (Lower is better) values are reported as the mean $\pm$ standard deviation across the test set.}
    \label{tab:table1}
   
    \resizebox{\textwidth}{!}{ 
        \begin{tabular}{c c c c c c c c c c} 
            \toprule
            & \multicolumn{3}{c}{Pelvic Dataset} & \multicolumn{3}{c}{Head Dataset} & \multicolumn{3}{c}{SynthRAD2023 (Brain)} \\
            \cmidrule{2-4} \cmidrule{5-7} \cmidrule{8-10} 
            Method & PSNR $\uparrow$ & SSIM $\uparrow$ & LPIPS $\downarrow$ & PSNR $\uparrow$ & SSIM $\uparrow$ & LPIPS $\downarrow$ & PSNR $\uparrow$ & SSIM $\uparrow$ & LPIPS $\downarrow$ \\
            \midrule
            CycleGAN  & 25.99 $\pm$ 2.03 & 83.15 $\pm$ 2.03 & 0.1155 $\pm$ 0.030 & 21.55 $\pm$ 1.38 & 82.27 $\pm$ 4.63 & 0.1315 $\pm$ 0.019 & 17.26 $\pm$ 1.11 & 56.04 $\pm$ 5.05 & 0.3078 $\pm$ 0.020 \\
            UNIT  & 23.04 $\pm$ 2.64 & 88.85 $\pm$ 3.00 & 0.1142 $\pm$ 0.021 & 23.06 $\pm$ 2.02 & 85.55 $\pm$ 4.17 & 0.1346 $\pm$ 0.025 & 24.02 $\pm$ 2.43 & 84.81 $\pm$ 5.31 & 0.1592 $\pm$ 0.025 \\
            CUT  & 26.30 $\pm$ 1.16 & 81.75 $\pm$ 2.63 & 0.1391 $\pm$ 0.018 & 16.83 $\pm$ 1.44 & 69.09 $\pm$ 5.85 & 0.2357 $\pm$ 0.025 & 15.43 $\pm$ 1.06 & 25.90 $\pm$ 6.01 & 0.4844 $\pm$ 0.014 \\
            SynDiff  & 26.02 $\pm$ 1.82 & 89.68 $\pm$ 2.32 & 0.0999 $\pm$ 0.016 & 22.90 $\pm$ 1.94 & 85.58 $\pm$ 5.10 & 0.1305 $\pm$ 0.021 & 24.97 $\pm$ 2.88 & 84.02 $\pm$ 4.34 & \textbf{0.1331 $\pm$ 0.025} \\
            Senta  & 27.73 $\pm$ 1.76 & 88.34 $\pm$ 2.44 & 0.1174 $\pm$ 0.018 & 22.93 $\pm$ 1.96 & 85.10 $\pm$ 4.99 & 0.1242 $\pm$ 0.022 & 24.05 $\pm$ 2.30 & 83.88 $\pm$ 5.34 & 0.1573 $\pm$ 0.023 \\
            Ours & \textbf{28.52 $\pm$ 1.77} & \textbf{90.56 $\pm$ 2.10} & \textbf{0.0915 $\pm$ 0.015} & \textbf{23.74 $\pm$ 2.11} & \textbf{86.72 $\pm$ 4.71} & \textbf{0.1162 $\pm$ 0.021} & \textbf{25.44 $\pm$ 2.73} & \textbf{86.17 $\pm$ 5.85} & 0.1365 $\pm$ 0.021 \\
            \bottomrule
        \end{tabular}
    } 
\end{table}

\begin{figure*}[!t]
    \centering
    \includegraphics[width=0.7\textwidth]{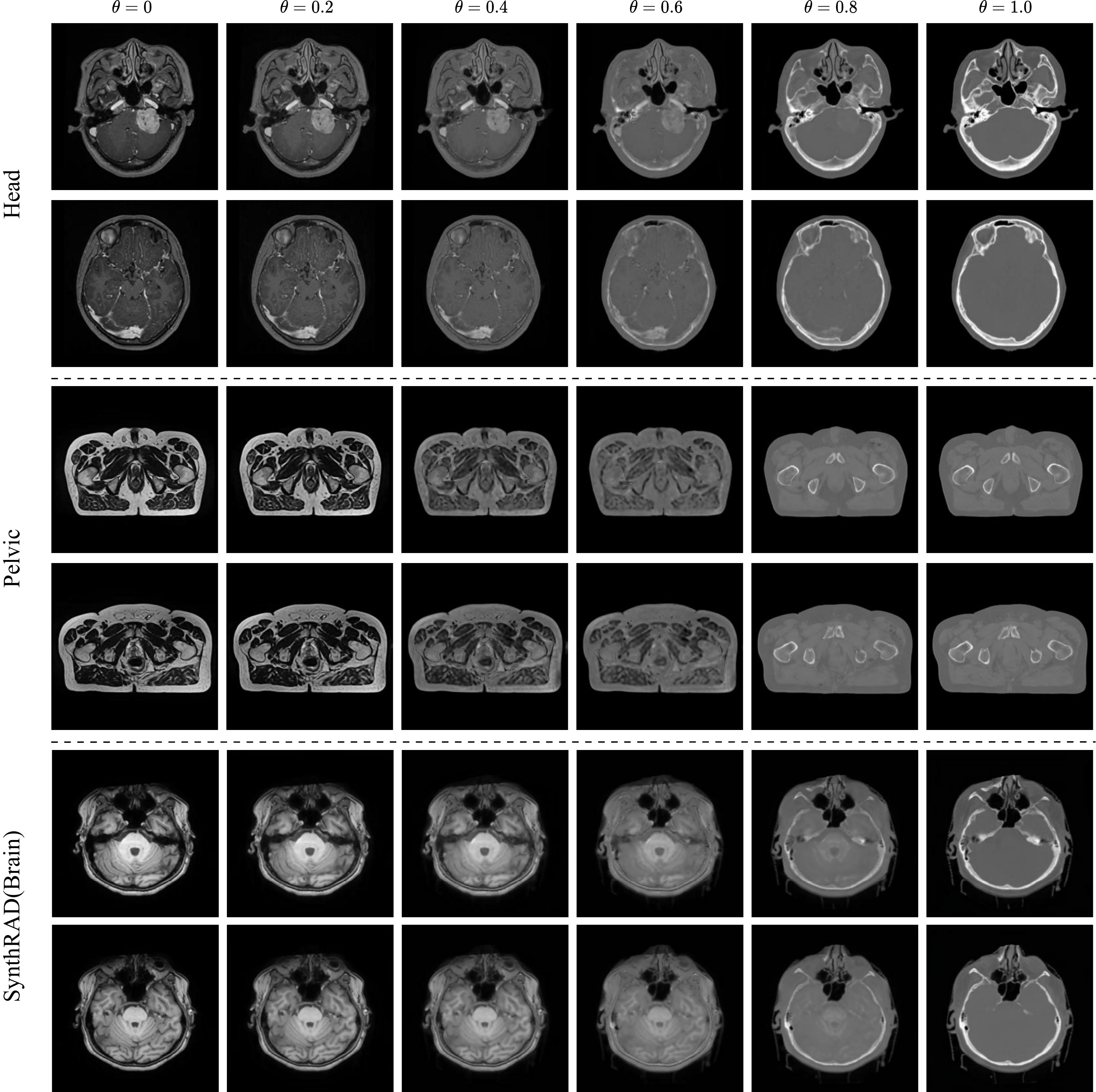}
    \caption{The proposed method translates MRI to CT in a continuous fashion over $\theta\in[0,1]$. From left to right, the MRI image gradually transitions to its corresponding CT representation. From top to bottom, two examples are presented for each of the Head, Pelvic, and SynthRAD datasets.}
    \label{fig6}
\end{figure*}

In the experiments, we used the Peak Signal-to-Noise Ratio (PSNR), Structural Similarity Index (SSIM), and Learned Perceptual Image Patch Similarity (LPIPS) \citep{LPIPS} as evaluation metrics. Higher values of PSNR and SSIM, along with lower values of LPIPS, indicate reduced distortion and superior translation quality. As summarized in Table~\ref{tab:table1}, our method consistently outperforms all competing approaches across all metrics and datasets, demonstrating its efficacy and robustness in unpaired MRI (both T1 and T2) to CT translation. We also provide a qualitative comparison of the translated CT images. As shown in Fig.~\ref{fig5}, our results demonstrate a closer visual similarity to the reference images, at the macro and micro structural levels. Fig.~\ref{fig6} further showcases the progressive transition of the MR images to the translated CT images at different time steps of the variable $\theta$.

\subsubsection{Bone Region Segmentation Results}

\begin{figure}[!t]
\centerline{\includegraphics[width=\columnwidth]{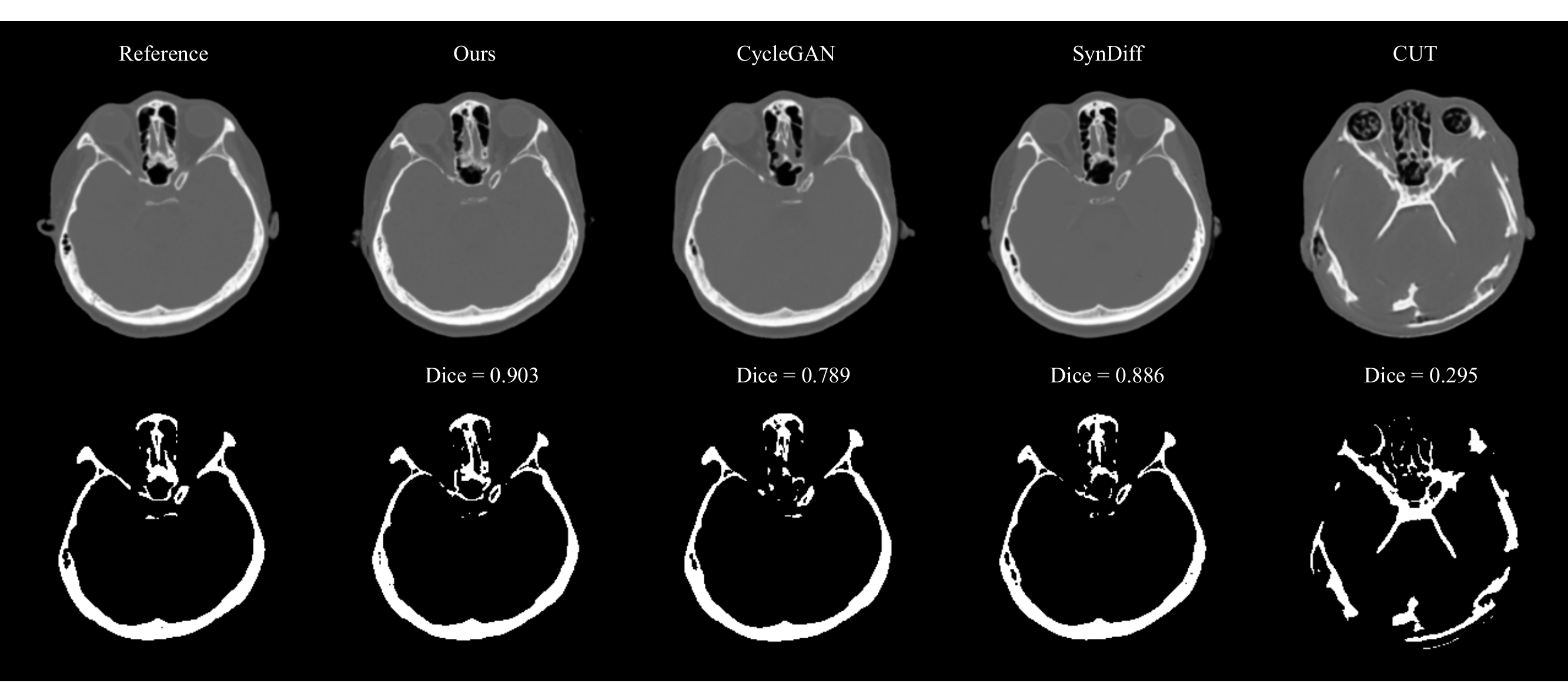}}
\caption{Visual comparison of bone segmentation results on the head dataset. The second row displays the segmented bones. From left to right: the reference CT, followed by translated CTs generated by our method, CycleGAN, SynDiff, and CUT. }
\label{fig7}
\end{figure}

To assess the efficacy of the proposed bone-contour regularization mechanism, we performed a downstream segmentation task specifically focused on bone regions using the head dataset. In this experiment, bones on CT were automatically segmented using a threshold-based approach, wherein voxels with Hounsfield unit (HU) values exceeding 300 were classified as bone tissue.

For quantitative evaluation, we calculated the Dice coefficient between the segmented bone regions in the reference CT and those in translated CT images. As reported in Table~\ref{tab:table2}, our method achieved a mean Dice coefficient of 0.84 ± 0.06, surpassing all competing approaches. These results underscore the effectiveness of our approach in maintaining the anatomical integrity of bone structures during MRI to CT translation, which is particularly advantageous for downstream clinical applications. Fig.~\ref{fig7} provides a qualitative visualization of the translated bone structures in the head dataset.

\begin{table}[h]
\centering
\caption{Dice Coefficients for Bone Segmentation on the Head Dataset}
\label{tab:table2}
\begin{tabular}{lc}
\toprule
Method & Dice Coefficient (mean ± std) \\ 
\midrule
CycleGAN & 0.76 ± 0.07 \\ 
UNIT  & 0.80 ± 0.07 \\ 
CUT  & 0.32 ± 0.13 \\ 
SynDiff  & 0.81 ± 0.07 \\ 
Senta & 0.81 ± 0.06 \\ 
Ours & \textbf{0.84 ± 0.06} \\ 
\bottomrule
\end{tabular}
\end{table}

\subsubsection{Ablation Studies}
We conducted a series of ablation studies to assess the contributions of key components in our model. 

We first present the ablation study evaluating the effects of path- and bone-contour regularization strategies. As shown in Table~\ref{tab:table3}, the baseline configuration (Setting A), which excludes any regularization, yielded suboptimal performance. Incorporating path regularization in Setting B led to a marked improvement, demonstrating its efficacy in MRI-to-CT translation. Further enhancement was observed in Setting C, where the addition of bone-contour regularization resulted in significant gains across all evaluation metrics, underscoring the complementary value of this strategy.

\begin{table}[htbp]
    \centering
    \caption{Quantitative Evaluation of Design Elements in Translation Performance.}
    \label{tab:table3}
    \begin{tabular}{@{}lcccccc@{}}
        \toprule
        & \multicolumn{2}{c}{Settings} & \multicolumn{2}{c}{Head Dataset} & \multicolumn{2}{c}{Pelvic Dataset} \\
        \cmidrule(lr){2-3} \cmidrule(lr){4-5} \cmidrule(lr){6-7}
        & Path & Contour & PSNR (dB) & SSIM & PSNR (dB) & SSIM \\
        \midrule
        A & -- & -- & 22.60$\pm$1.91 & 84.24$\pm$5.15 & 26.49$\pm$2.06 & 85.05$\pm$2.64 \\
        B & \checkmark & -- & 23.11$\pm$1.96 & 85.56$\pm$4.91 & 27.68$\pm$1.62 & 88.18$\pm$2.47 \\
        C & \checkmark & \checkmark & \textbf{23.74$\pm$2.11} & \textbf{86.72$\pm$4.71} & \textbf{28.52$\pm$1.77} & \textbf{90.56$\pm$2.10} \\
        \bottomrule
    \end{tabular}
\end{table}

Next, we evaluated the effectiveness of the multi-scale path regularization and the bone neighboring region attention map $W$. As presented in Table~\ref{tab:table4}, removing either component resulted in decreased performance on both the head and pelvic datasets, highlighting their contributions to the proposed approach.

\begin{table}[htbp]
    \centering
    \caption{Effect of the multi-scale path regularization and the bone neighboring region attention map}
    \label{tab:table4}
    \begin{tabular}{@{}lcccc@{}}
        \toprule 
        & \multicolumn{2}{c}{Head Dataset} & \multicolumn{2}{c}{Pelvic Dataset} \\
        \cmidrule(lr){2-3} \cmidrule(lr){4-5}
        & PSNR & SSIM & PSNR & SSIM \\
        \midrule
        Ours & \textbf{23.74$\pm$2.11} & \textbf{86.72$\pm$4.71} & \textbf{28.52$\pm$1.77} & \textbf{90.56$\pm$2.10} \\
        w/o multi-scale & 23.47$\pm$2.07 & 86.63$\pm$4.75 & 28.34$\pm$1.60 & 89.23$\pm$2.01 \\
        w/o attention map $W$ & 23.29$\pm$2.02 & 86.15$\pm$4.64 & 27.65$\pm$1.99 & 89.81$\pm$2.22 \\
        \bottomrule
    \end{tabular}
\end{table}

Finally, we assessed the contribution of the proposed trainable bone-contour generator $G_\mathrm{bone}$. Specifically, we compared three scenarios: (1) using the proposed bone-contour loss $\mathcal{L_\mathrm{bone}}$, computed with $G_\mathrm{bone}$; (2) replacing the $G_\mathrm{bone}(x)$ term in $\mathcal{L_\mathrm{bone}}$ with edge information extracted from MRI using a conventional Sobel filter; and (3) excluding $\mathcal{L_\mathrm{bone}}$ entirely. As shown in Table~\ref{tab:table5}, the Sobel-based approach resulted in inferior performance compared to the proposed method, likely due to the difficulty of accurately capturing bone contours in MR images using handcrafted filters. These results highlight the importance of incorporating a trainable bone-contour generator into the model to learn and infer plausible CT contours from MRI inputs.

\begin{table}[htbp]
    \centering
    \caption{Impact of Contour Information Strategies on Image Generation Quality}
    \label{tab:table5}
    \begin{tabular}{@{}lcccc@{}}
        \toprule
        & \multicolumn{2}{c}{Head} & \multicolumn{2}{c}{Pelvic} \\
        \cmidrule(lr){2-3} \cmidrule(lr){4-5}
        Method & PSNR & SSIM & PSNR & SSIM \\
        \midrule
        Ours (Contour) & \textbf{23.74$\pm$2.11} & \textbf{86.72$\pm$4.71} & \textbf{28.52$\pm$1.77} & \textbf{90.56$\pm$2.10} \\
        MRI Contours   & 23.23$\pm$1.91 & 85.89$\pm$4.54 & 28.00$\pm$1.85 & 88.32$\pm$2.24 \\
        No Contours    & 23.11$\pm$1.96 & 85.56$\pm$4.91 & 27.68$\pm$1.62 & 88.18$\pm$2.47 \\
        \bottomrule
    \end{tabular}
\end{table}

\section{Discussion and Conclusion}

Unpaired MRI-to-CT translation is of significant clinical importance, as it enables clinicians to harness the complementary strengths of both imaging modalities without the need for additional CT scans. Our approach employs a neural ordinary differential equation (ODE) framework, formulating MRI-to-CT translation as a latent space path transition problem. By minimizing the length of the transition path, the model achieves a more efficient solution and demonstrates superior performance compared to conventional cycle-consistency and contrastive learning methods. The introduction of a trainable bone-contour generator represents a notable advancement over traditional filter-based methods, as it learns to infer plausible bone contours directly from MRI data and enforces their preservation in the translated CT images. Ablation studies further validate the complementary effect of integrating path regularization with bone-contour regularization, demonstrating their combined contribution to improved translation performance. The proposed mechanism for preserving the structural fidelity of translated bone anatomy is essential, particularly given the critical role of bones in radiation therapy. This enhancement positions our approach as a highly relevant tool for clinical application in radiotherapy. Moreover, the mechanism is extensible to other anatomically significant structures, accommodating variations in clinical priorities across different tasks. By enabling the targeted preservation of specific structures, our method offers a flexible and generalizable framework suited to a wide range of diagnostic and therapeutic applications.
In future works, we will focus on validate the proposed approach in clinical settings, particularly through trials involving radiotherapy dose planning. Such validation would offer concrete evidence of the method’s clinical utility and its potential to improve treatment outcomes.

\printcredits

\section*{Declaration of competing interest}
The authors declare that they have no known competing financial 
interests or personal relationships that could have appeared to influence 
the work reported in this paper.

\section*{Acknowledgments}
This research was supported in part by the
Basic and Applied Basic Research Foundation of Guangdong Province (2025A1515012208, 2023A1515011644, 2024A1515011697), and Guangdong Province Key Technologies R\&D Program for ``Brain Science and Brain-like Intelligence Research'' (2023B0303020002).

\bibliographystyle{cas-model2-names}

\bibliography{cas-refs}

\begin{thebibliography}{40}
\expandafter\ifx\csname natexlab\endcsname\relax\def\natexlab#1{#1}\fi
\providecommand{\url}[1]{\texttt{#1}}
\providecommand{\href}[2]{#2}
\providecommand{\path}[1]{#1}
\providecommand{\DOIprefix}{doi:}
\providecommand{\ArXivprefix}{arXiv:}
\providecommand{\URLprefix}{URL: }
\providecommand{\Pubmedprefix}{pmid:}
\providecommand{\doi}[1]{\href{http://dx.doi.org/#1}{\path{#1}}}
\providecommand{\Pubmed}[1]{\href{pmid:#1}{\path{#1}}}
\providecommand{\bibinfo}[2]{#2}
\ifx\xfnm\relax \def\xfnm[#1]{\unskip,\space#1}\fi
\bibitem[{Arslan et~al.(2024)Arslan, Kabas, Dalmaz, Ozbey and {\c{C}}ukur}]{Bridge1}
\bibinfo{author}{Arslan, F.}, \bibinfo{author}{Kabas, B.}, \bibinfo{author}{Dalmaz, O.}, \bibinfo{author}{Ozbey, M.}, \bibinfo{author}{{\c{C}}ukur, T.}, \bibinfo{year}{2024}.
\newblock \bibinfo{title}{Self-consistent recursive diffusion bridge for medical image translation}.
\newblock \bibinfo{journal}{arXiv preprint arXiv:2405.06789} .
\bibitem[{Boni et~al.(2020)Boni, Klein, Vanquin, Wagner, Lacornerie, Pasquier and Reynaert}]{cGAN3}
\bibinfo{author}{Boni, K.N.B.}, \bibinfo{author}{Klein, J.}, \bibinfo{author}{Vanquin, L.}, \bibinfo{author}{Wagner, A.}, \bibinfo{author}{Lacornerie, T.}, \bibinfo{author}{Pasquier, D.}, \bibinfo{author}{Reynaert, N.}, \bibinfo{year}{2020}.
\newblock \bibinfo{title}{Mr to ct synthesis with multicenter data in the pelvic area using a conditional generative adversarial network}.
\newblock \bibinfo{journal}{Physics in Medicine \& Biology} \bibinfo{volume}{65}, \bibinfo{pages}{075002}.
\bibitem[{Burgos et~al.(2015)Burgos, Cardoso, Guerreiro, Veiga, Modat, McClelland, Knopf, Punwani, Atkinson, Arridge et~al.}]{burgos2015robust}
\bibinfo{author}{Burgos, N.}, \bibinfo{author}{Cardoso, M.J.}, \bibinfo{author}{Guerreiro, F.}, \bibinfo{author}{Veiga, C.}, \bibinfo{author}{Modat, M.}, \bibinfo{author}{McClelland, J.}, \bibinfo{author}{Knopf, A.C.}, \bibinfo{author}{Punwani, S.}, \bibinfo{author}{Atkinson, D.}, \bibinfo{author}{Arridge, S.R.}, et~al., \bibinfo{year}{2015}.
\newblock \bibinfo{title}{Robust ct synthesis for radiotherapy planning: application to the head and neck region}, in: \bibinfo{booktitle}{Medical Image Computing and Computer-Assisted Intervention--MICCAI 2015: 18th International Conference, Munich, Germany, October 5-9, 2015, Proceedings, Part II 18}, \bibinfo{organization}{Springer}. pp. \bibinfo{pages}{476--484}.
\bibitem[{Cai et~al.(2019)Cai, Zhang, Cui, Zheng and Yang}]{CycleGAN3}
\bibinfo{author}{Cai, J.}, \bibinfo{author}{Zhang, Z.}, \bibinfo{author}{Cui, L.}, \bibinfo{author}{Zheng, Y.}, \bibinfo{author}{Yang, L.}, \bibinfo{year}{2019}.
\newblock \bibinfo{title}{Towards cross-modal organ translation and segmentation: A cycle-and shape-consistent generative adversarial network}.
\newblock \bibinfo{journal}{Medical image analysis} \bibinfo{volume}{52}, \bibinfo{pages}{174--184}.
\bibitem[{Chen et~al.(2024)Chen, Konz, Gu, Dong, Chen, Li, Lee and Mazurowski}]{Contour_diff}
\bibinfo{author}{Chen, Y.}, \bibinfo{author}{Konz, N.}, \bibinfo{author}{Gu, H.}, \bibinfo{author}{Dong, H.}, \bibinfo{author}{Chen, Y.}, \bibinfo{author}{Li, L.}, \bibinfo{author}{Lee, J.}, \bibinfo{author}{Mazurowski, M.A.}, \bibinfo{year}{2024}.
\newblock \bibinfo{title}{Contourdiff: Unpaired image translation with contour-guided diffusion models}.
\newblock \bibinfo{journal}{arXiv preprint arXiv:2403.10786} .
\bibitem[{Chen et~al.(2025)Chen, Wang, Lin, Peng, Liu, Luo, Wang, Liu, Haouchine and Qiao}]{DDPM2025}
\bibinfo{author}{Chen, Z.}, \bibinfo{author}{Wang, L.}, \bibinfo{author}{Lin, Y.}, \bibinfo{author}{Peng, J.}, \bibinfo{author}{Liu, Z.}, \bibinfo{author}{Luo, J.}, \bibinfo{author}{Wang, B.}, \bibinfo{author}{Liu, Y.}, \bibinfo{author}{Haouchine, N.}, \bibinfo{author}{Qiao, X.}, \bibinfo{year}{2025}.
\newblock \bibinfo{title}{Backward stochastic differential equations-guided generative model for structural-to-functional neuroimage translator}.
\newblock \bibinfo{journal}{arXiv} .
\bibitem[{Chen et~al.(2022)Chen, Wei and Li}]{CUT1}
\bibinfo{author}{Chen, Z.}, \bibinfo{author}{Wei, J.}, \bibinfo{author}{Li, R.}, \bibinfo{year}{2022}.
\newblock \bibinfo{title}{Unsupervised multi-modal medical image registration via discriminator-free image-to-image translation}.
\newblock \bibinfo{journal}{arXiv preprint arXiv:2204.13656} .
\bibitem[{Choi et~al.(2023)Choi, Jeong, Lee, Lee and Kim}]{CUT4}
\bibinfo{author}{Choi, C.}, \bibinfo{author}{Jeong, J.}, \bibinfo{author}{Lee, S.}, \bibinfo{author}{Lee, S.M.}, \bibinfo{author}{Kim, N.}, \bibinfo{year}{2023}.
\newblock \bibinfo{title}{Ct kernel conversion using multi-domain image-to-image translation with generator-guided contrastive learning}, in: \bibinfo{booktitle}{International Conference on Medical Image Computing and Computer-Assisted Intervention}, \bibinfo{organization}{Springer}. pp. \bibinfo{pages}{344--354}.
\bibitem[{Dalmaz et~al.(2022)Dalmaz, Yurt and {\c{C}}ukur}]{ResViT}
\bibinfo{author}{Dalmaz, O.}, \bibinfo{author}{Yurt, M.}, \bibinfo{author}{{\c{C}}ukur, T.}, \bibinfo{year}{2022}.
\newblock \bibinfo{title}{Resvit: residual vision transformers for multimodal medical image synthesis}.
\newblock \bibinfo{journal}{IEEE Transactions on Medical Imaging} \bibinfo{volume}{41}, \bibinfo{pages}{2598--2614}.
\bibitem[{Dayarathna et~al.(2023)Dayarathna, Islam, Uribe, Yang, Hayat and Chen}]{review1}
\bibinfo{author}{Dayarathna, S.}, \bibinfo{author}{Islam, K.T.}, \bibinfo{author}{Uribe, S.}, \bibinfo{author}{Yang, G.}, \bibinfo{author}{Hayat, M.}, \bibinfo{author}{Chen, Z.}, \bibinfo{year}{2023}.
\newblock \bibinfo{title}{Deep learning based synthesis of mri, ct and pet: Review and analysis}.
\newblock \bibinfo{journal}{Medical Image Analysis} , \bibinfo{pages}{103046}.
\bibitem[{Dhariwal and Nichol(2021)}]{DDPM_theory1}
\bibinfo{author}{Dhariwal, P.}, \bibinfo{author}{Nichol, A.}, \bibinfo{year}{2021}.
\newblock \bibinfo{title}{Diffusion models beat gans on image synthesis}.
\newblock \bibinfo{journal}{Advances in neural information processing systems} \bibinfo{volume}{34}, \bibinfo{pages}{8780--8794}.
\bibitem[{Emami et~al.(2018)Emami, Dong, Nejad-Davarani and Glide-Hurst}]{cGAN1}
\bibinfo{author}{Emami, H.}, \bibinfo{author}{Dong, M.}, \bibinfo{author}{Nejad-Davarani, S.P.}, \bibinfo{author}{Glide-Hurst, C.K.}, \bibinfo{year}{2018}.
\newblock \bibinfo{title}{Generating synthetic cts from magnetic resonance images using generative adversarial networks}.
\newblock \bibinfo{journal}{Medical physics} \bibinfo{volume}{45}, \bibinfo{pages}{3627--3636}.
\bibitem[{Ge et~al.(2019)Ge, Wei, Xue, Wang, Zhou, Zhan and Liao}]{CycleGAN5}
\bibinfo{author}{Ge, Y.}, \bibinfo{author}{Wei, D.}, \bibinfo{author}{Xue, Z.}, \bibinfo{author}{Wang, Q.}, \bibinfo{author}{Zhou, X.}, \bibinfo{author}{Zhan, Y.}, \bibinfo{author}{Liao, S.}, \bibinfo{year}{2019}.
\newblock \bibinfo{title}{Unpaired mr to ct synthesis with explicit structural constrained adversarial learning}, in: \bibinfo{booktitle}{2019 IEEE 16th International Symposium on Biomedical Imaging (ISBI 2019)}, \bibinfo{organization}{IEEE}. pp. \bibinfo{pages}{1096--1099}.
\bibitem[{Goodfellow et~al.(2014)Goodfellow, Pouget-Abadie, Mirza, Xu, Warde-Farley, Ozair, Courville and Bengio}]{GAN_theory1}
\bibinfo{author}{Goodfellow, I.}, \bibinfo{author}{Pouget-Abadie, J.}, \bibinfo{author}{Mirza, M.}, \bibinfo{author}{Xu, B.}, \bibinfo{author}{Warde-Farley, D.}, \bibinfo{author}{Ozair, S.}, \bibinfo{author}{Courville, A.}, \bibinfo{author}{Bengio, Y.}, \bibinfo{year}{2014}.
\newblock \bibinfo{title}{Generative adversarial nets}.
\newblock \bibinfo{journal}{Advances in neural information processing systems} \bibinfo{volume}{27}.
\bibitem[{Ho et~al.(2020)Ho, Jain and Abbeel}]{DDPM_theory2}
\bibinfo{author}{Ho, J.}, \bibinfo{author}{Jain, A.}, \bibinfo{author}{Abbeel, P.}, \bibinfo{year}{2020}.
\newblock \bibinfo{title}{Denoising diffusion probabilistic models}.
\newblock \bibinfo{journal}{Advances in neural information processing systems} \bibinfo{volume}{33}, \bibinfo{pages}{6840--6851}.
\bibitem[{Hognon et~al.(2024)Hognon, Conze, Bourbonne, Gallinato, Colin, Jaouen and Visvikis}]{CUT2}
\bibinfo{author}{Hognon, C.}, \bibinfo{author}{Conze, P.H.}, \bibinfo{author}{Bourbonne, V.}, \bibinfo{author}{Gallinato, O.}, \bibinfo{author}{Colin, T.}, \bibinfo{author}{Jaouen, V.}, \bibinfo{author}{Visvikis, D.}, \bibinfo{year}{2024}.
\newblock \bibinfo{title}{Contrastive image adaptation for acquisition shift reduction in medical imaging}.
\newblock \bibinfo{journal}{Artificial Intelligence in Medicine} \bibinfo{volume}{148}, \bibinfo{pages}{102747}.
\bibitem[{Hu et~al.(2024)Hu, Yoon, Wu, Tivnan, Chen, Wang, Luo, Cui, Li, Liu and Guo}]{cDDPM3}
\bibinfo{author}{Hu, R.}, \bibinfo{author}{Yoon, S.}, \bibinfo{author}{Wu, D.}, \bibinfo{author}{Tivnan, M.}, \bibinfo{author}{Chen, Z.}, \bibinfo{author}{Wang, Y.}, \bibinfo{author}{Luo, J.}, \bibinfo{author}{Cui, J.}, \bibinfo{author}{Li, Q.}, \bibinfo{author}{Liu, H.}, \bibinfo{author}{Guo, N.}, \bibinfo{year}{2024}.
\newblock \bibinfo{title}{Realistic tumor generation using 3d conditional latent diffusion model}.
\newblock \bibinfo{journal}{Journal of Nuclear Medicine} \bibinfo{volume}{65}.
\bibitem[{Isola et~al.(2017)Isola, Zhu, Zhou and Efros}]{pix2pix}
\bibinfo{author}{Isola, P.}, \bibinfo{author}{Zhu, J.Y.}, \bibinfo{author}{Zhou, T.}, \bibinfo{author}{Efros, A.A.}, \bibinfo{year}{2017}.
\newblock \bibinfo{title}{Image-to-image translation with conditional adversarial networks}, in: \bibinfo{booktitle}{Proceedings of the IEEE conference on computer vision and pattern recognition}, pp. \bibinfo{pages}{1125--1134}.
\bibitem[{Kazerouni et~al.(2023)Kazerouni, Aghdam, Heidari, Azad, Fayyaz, Hacihaliloglu and Merhof}]{review2}
\bibinfo{author}{Kazerouni, A.}, \bibinfo{author}{Aghdam, E.K.}, \bibinfo{author}{Heidari, M.}, \bibinfo{author}{Azad, R.}, \bibinfo{author}{Fayyaz, M.}, \bibinfo{author}{Hacihaliloglu, I.}, \bibinfo{author}{Merhof, D.}, \bibinfo{year}{2023}.
\newblock \bibinfo{title}{Diffusion models in medical imaging: A comprehensive survey}.
\newblock \bibinfo{journal}{Medical Image Analysis} , \bibinfo{pages}{102846}.
\bibitem[{Kearney et~al.(2020)Kearney, Ziemer, Perry, Wang, Chan, Ma, Morin, Yom and Solberg}]{CycleGAN2}
\bibinfo{author}{Kearney, V.}, \bibinfo{author}{Ziemer, B.P.}, \bibinfo{author}{Perry, A.}, \bibinfo{author}{Wang, T.}, \bibinfo{author}{Chan, J.W.}, \bibinfo{author}{Ma, L.}, \bibinfo{author}{Morin, O.}, \bibinfo{author}{Yom, S.S.}, \bibinfo{author}{Solberg, T.D.}, \bibinfo{year}{2020}.
\newblock \bibinfo{title}{Attention-aware discrimination for mr-to-ct image translation using cycle-consistent generative adversarial networks}.
\newblock \bibinfo{journal}{Radiology: Artificial Intelligence} \bibinfo{volume}{2}, \bibinfo{pages}{e190027}.
\bibitem[{Kim and Park(2024)}]{cDDPM-latent}
\bibinfo{author}{Kim, J.}, \bibinfo{author}{Park, H.}, \bibinfo{year}{2024}.
\newblock \bibinfo{title}{Adaptive latent diffusion model for 3d medical image to image translation: Multi-modal magnetic resonance imaging study}, in: \bibinfo{booktitle}{Proceedings of the IEEE/CVF Winter Conference on Applications of Computer Vision}, pp. \bibinfo{pages}{7604--7613}.
\bibitem[{Kingma et~al.(2013)Kingma, Welling et~al.}]{VAE}
\bibinfo{author}{Kingma, D.P.}, \bibinfo{author}{Welling, M.}, et~al., \bibinfo{year}{2013}.
\newblock \bibinfo{title}{Auto-encoding variational bayes}.
\bibitem[{Li et~al.(2023)Li, Shao, Qian and Zhang}]{FDDM}
\bibinfo{author}{Li, Y.}, \bibinfo{author}{Shao, H.C.}, \bibinfo{author}{Qian, X.}, \bibinfo{author}{Zhang, Y.}, \bibinfo{year}{2023}.
\newblock \bibinfo{title}{Fddm: Unsupervised medical image translation with a frequency-decoupled diffusion model}.
\newblock \bibinfo{journal}{Machine Learning: Science and Technology} .
\bibitem[{Liu et~al.(2017)Liu, Breuel and Kautz}]{UNIT_theory}
\bibinfo{author}{Liu, M.Y.}, \bibinfo{author}{Breuel, T.}, \bibinfo{author}{Kautz, J.}, \bibinfo{year}{2017}.
\newblock \bibinfo{title}{Unsupervised image-to-image translation networks}.
\newblock \bibinfo{journal}{Advances in neural information processing systems} \bibinfo{volume}{30}.
\bibitem[{Luo et~al.(2024)Luo, Zhang, Ling, Lin, Wang and Yao}]{cGAN4}
\bibinfo{author}{Luo, Y.}, \bibinfo{author}{Zhang, S.}, \bibinfo{author}{Ling, J.}, \bibinfo{author}{Lin, Z.}, \bibinfo{author}{Wang, Z.}, \bibinfo{author}{Yao, S.}, \bibinfo{year}{2024}.
\newblock \bibinfo{title}{Mask-guided generative adversarial network for mri-based ct synthesis}.
\newblock \bibinfo{journal}{Knowledge-Based Systems} \bibinfo{volume}{295}, \bibinfo{pages}{111799}.
\bibitem[{Lyu and Wang(2022)}]{cDDPM1}
\bibinfo{author}{Lyu, Q.}, \bibinfo{author}{Wang, G.}, \bibinfo{year}{2022}.
\newblock \bibinfo{title}{Conversion between ct and mri images using diffusion and score-matching models}.
\newblock \bibinfo{journal}{arXiv preprint arXiv:2209.12104} .
\bibitem[{Nyholm et~al.(2018)Nyholm, Svensson, Andersson, Jonsson, Sohlin, Gustafsson, Kjell{\'e}n, S{\"o}derstr{\"o}m, Albertsson, Blomqvist et~al.}]{nyholm2018mr}
\bibinfo{author}{Nyholm, T.}, \bibinfo{author}{Svensson, S.}, \bibinfo{author}{Andersson, S.}, \bibinfo{author}{Jonsson, J.}, \bibinfo{author}{Sohlin, M.}, \bibinfo{author}{Gustafsson, C.}, \bibinfo{author}{Kjell{\'e}n, E.}, \bibinfo{author}{S{\"o}derstr{\"o}m, K.}, \bibinfo{author}{Albertsson, P.}, \bibinfo{author}{Blomqvist, L.}, et~al., \bibinfo{year}{2018}.
\newblock \bibinfo{title}{Mr and ct data with multiobserver delineations of organs in the pelvic area—part of the gold atlas project}.
\newblock \bibinfo{journal}{Medical physics} \bibinfo{volume}{45}, \bibinfo{pages}{1295--1300}.
\bibitem[{{\"O}zbey et~al.(2023){\"O}zbey, Dalmaz, Dar, Bedel, {\"O}zturk, G{\"u}ng{\"o}r and {\c{C}}ukur}]{SynDiff}
\bibinfo{author}{{\"O}zbey, M.}, \bibinfo{author}{Dalmaz, O.}, \bibinfo{author}{Dar, S.U.}, \bibinfo{author}{Bedel, H.A.}, \bibinfo{author}{{\"O}zturk, {\c{S}}.}, \bibinfo{author}{G{\"u}ng{\"o}r, A.}, \bibinfo{author}{{\c{C}}ukur, T.}, \bibinfo{year}{2023}.
\newblock \bibinfo{title}{Unsupervised medical image translation with adversarial diffusion models}.
\newblock \bibinfo{journal}{IEEE Transactions on Medical Imaging} .
\bibitem[{Pan et~al.(2024)Pan, Abouei, Wynne, Chang, Wang, Qiu, Li, Peng, Roper, Patel et~al.}]{cDDPM2}
\bibinfo{author}{Pan, S.}, \bibinfo{author}{Abouei, E.}, \bibinfo{author}{Wynne, J.}, \bibinfo{author}{Chang, C.W.}, \bibinfo{author}{Wang, T.}, \bibinfo{author}{Qiu, R.L.}, \bibinfo{author}{Li, Y.}, \bibinfo{author}{Peng, J.}, \bibinfo{author}{Roper, J.}, \bibinfo{author}{Patel, P.}, et~al., \bibinfo{year}{2024}.
\newblock \bibinfo{title}{Synthetic ct generation from mri using 3d transformer-based denoising diffusion model}.
\newblock \bibinfo{journal}{Medical Physics} \bibinfo{volume}{51}, \bibinfo{pages}{2538--2548}.
\bibitem[{Park et~al.(2020)Park, Efros, Zhang and Zhu}]{CUT_theory}
\bibinfo{author}{Park, T.}, \bibinfo{author}{Efros, A.A.}, \bibinfo{author}{Zhang, R.}, \bibinfo{author}{Zhu, J.Y.}, \bibinfo{year}{2020}.
\newblock \bibinfo{title}{Contrastive learning for unpaired image-to-image translation}, in: \bibinfo{booktitle}{Computer Vision--ECCV 2020: 16th European Conference, Glasgow, UK, August 23--28, 2020, Proceedings, Part IX 16}, \bibinfo{organization}{Springer}. pp. \bibinfo{pages}{319--345}.
\bibitem[{Sj{\"o}lund et~al.(2015)Sj{\"o}lund, Forsberg, Andersson and Knutsson}]{sjolund2015generating}
\bibinfo{author}{Sj{\"o}lund, J.}, \bibinfo{author}{Forsberg, D.}, \bibinfo{author}{Andersson, M.}, \bibinfo{author}{Knutsson, H.}, \bibinfo{year}{2015}.
\newblock \bibinfo{title}{Generating patient specific pseudo-ct of the head from mr using atlas-based regression}.
\newblock \bibinfo{journal}{Physics in Medicine \& Biology} \bibinfo{volume}{60}, \bibinfo{pages}{825}.
\bibitem[{Tan et~al.(2024)Tan, Patel, Wang, Luo, Chen, Luo, Chen, Mao, Huang, Wang et~al.}]{cGAN2}
\bibinfo{author}{Tan, Y.}, \bibinfo{author}{Patel, R.V.}, \bibinfo{author}{Wang, Z.}, \bibinfo{author}{Luo, Y.}, \bibinfo{author}{Chen, J.}, \bibinfo{author}{Luo, J.}, \bibinfo{author}{Chen, W.}, \bibinfo{author}{Mao, Z.}, \bibinfo{author}{Huang, R.Y.}, \bibinfo{author}{Wang, H.}, et~al., \bibinfo{year}{2024}.
\newblock \bibinfo{title}{Generation and applications of synthetic computed tomography images for neurosurgical planning}.
\newblock \bibinfo{journal}{Journal of Neurosurgery} \bibinfo{volume}{1}, \bibinfo{pages}{1--10}.
\bibitem[{Thummerer et~al.(2023)Thummerer, van~der Bijl, Galapon~Jr, Verhoeff, Langendijk, Both, van~den Berg and Maspero}]{thummerer2023synthrad2023}
\bibinfo{author}{Thummerer, A.}, \bibinfo{author}{van~der Bijl, E.}, \bibinfo{author}{Galapon~Jr, A.}, \bibinfo{author}{Verhoeff, J.J.}, \bibinfo{author}{Langendijk, J.A.}, \bibinfo{author}{Both, S.}, \bibinfo{author}{van~den Berg, C.N.A.}, \bibinfo{author}{Maspero, M.}, \bibinfo{year}{2023}.
\newblock \bibinfo{title}{Synthrad2023 grand challenge dataset: Generating synthetic ct for radiotherapy}.
\newblock \bibinfo{journal}{Medical physics} \bibinfo{volume}{50}, \bibinfo{pages}{4664--4674}.
\bibitem[{Wolterink et~al.(2017)Wolterink, Dinkla, Savenije, Seevinck, van~den Berg and I{\v{s}}gum}]{CycleGAN1}
\bibinfo{author}{Wolterink, J.M.}, \bibinfo{author}{Dinkla, A.M.}, \bibinfo{author}{Savenije, M.H.F.}, \bibinfo{author}{Seevinck, P.R.}, \bibinfo{author}{van~den Berg, C.A.T.}, \bibinfo{author}{I{\v{s}}gum, I.}, \bibinfo{year}{2017}.
\newblock \bibinfo{title}{Deep mr to ct synthesis using unpaired data}, in: \bibinfo{editor}{Tsaftaris, S.A.}, \bibinfo{editor}{Gooya, A.}, \bibinfo{editor}{Frangi, A.F.}, \bibinfo{editor}{Prince, J.L.} (Eds.), \bibinfo{booktitle}{Simulation and Synthesis in Medical Imaging}, \bibinfo{publisher}{Springer International Publishing}, \bibinfo{address}{Cham}. pp. \bibinfo{pages}{14--23}.
\bibitem[{Xie et~al.(2023)Xie, Xu, Gong and Zhang}]{Senta}
\bibinfo{author}{Xie, S.}, \bibinfo{author}{Xu, Y.}, \bibinfo{author}{Gong, M.}, \bibinfo{author}{Zhang, K.}, \bibinfo{year}{2023}.
\newblock \bibinfo{title}{Unpaired image-to-image translation with shortest path regularization}, in: \bibinfo{booktitle}{Proceedings of the IEEE/CVF Conference on Computer Vision and Pattern Recognition}, pp. \bibinfo{pages}{10177--10187}.
\bibitem[{Yu et~al.(2019)Yu, Zhou, Wang, Shi, Fripp and Bourgeat}]{Ea-GANs}
\bibinfo{author}{Yu, B.}, \bibinfo{author}{Zhou, L.}, \bibinfo{author}{Wang, L.}, \bibinfo{author}{Shi, Y.}, \bibinfo{author}{Fripp, J.}, \bibinfo{author}{Bourgeat, P.}, \bibinfo{year}{2019}.
\newblock \bibinfo{title}{Ea-gans: edge-aware generative adversarial networks for cross-modality mr image synthesis}.
\newblock \bibinfo{journal}{IEEE transactions on medical imaging} \bibinfo{volume}{38}, \bibinfo{pages}{1750--1762}.
\bibitem[{Zhang et~al.(2018a)Zhang, Isola, Efros, Shechtman and Wang}]{LPIPS}
\bibinfo{author}{Zhang, R.}, \bibinfo{author}{Isola, P.}, \bibinfo{author}{Efros, A.A.}, \bibinfo{author}{Shechtman, E.}, \bibinfo{author}{Wang, O.}, \bibinfo{year}{2018}a.
\newblock \bibinfo{title}{The unreasonable effectiveness of deep features as a perceptual metric}, in: \bibinfo{booktitle}{Proceedings of the IEEE conference on computer vision and pattern recognition}, pp. \bibinfo{pages}{586--595}.
\bibitem[{Zhang et~al.(2018b)Zhang, Yang and Zheng}]{CycleGAN4}
\bibinfo{author}{Zhang, Z.}, \bibinfo{author}{Yang, L.}, \bibinfo{author}{Zheng, Y.}, \bibinfo{year}{2018}b.
\newblock \bibinfo{title}{Translating and segmenting multimodal medical volumes with cycle-and shape-consistency generative adversarial network}, in: \bibinfo{booktitle}{Proceedings of the IEEE conference on computer vision and pattern Recognition}, pp. \bibinfo{pages}{9242--9251}.
\bibitem[{Zhong et~al.(2023)Zhong, Chen, Shu, Zheng, Li, Chen, Wu, Ma, Feng and Yang}]{cGAN6}
\bibinfo{author}{Zhong, L.}, \bibinfo{author}{Chen, Z.}, \bibinfo{author}{Shu, H.}, \bibinfo{author}{Zheng, K.}, \bibinfo{author}{Li, Y.}, \bibinfo{author}{Chen, W.}, \bibinfo{author}{Wu, Y.}, \bibinfo{author}{Ma, J.}, \bibinfo{author}{Feng, Q.}, \bibinfo{author}{Yang, W.}, \bibinfo{year}{2023}.
\newblock \bibinfo{title}{Multi-scale tokens-aware transformer network for multi-region and multi-sequence mr-to-ct synthesis in a single model}.
\newblock \bibinfo{journal}{IEEE transactions on medical imaging} \bibinfo{volume}{43}, \bibinfo{pages}{794--806}.
\bibitem[{Zhu et~al.(2017)Zhu, Park, Isola and Efros}]{CycleGAN_theory}
\bibinfo{author}{Zhu, J.Y.}, \bibinfo{author}{Park, T.}, \bibinfo{author}{Isola, P.}, \bibinfo{author}{Efros, A.A.}, \bibinfo{year}{2017}.
\newblock \bibinfo{title}{Unpaired image-to-image translation using cycle-consistent adversarial networks}, in: \bibinfo{booktitle}{Proceedings of the IEEE international conference on computer vision}, pp. \bibinfo{pages}{2223--2232}.

\end{thebibliography}
\end{document}